%% file: ICAR2023.tex
\documentclass[letterpaper, 10 pt, conference]{ieeeconf}
\IEEEoverridecommandlockouts
\usepackage{cite}
\usepackage{amsmath,amssymb,amsfonts}
\usepackage{algorithmic}
\usepackage{graphicx}
\usepackage{textcomp}
\usepackage{xcolor}
\usepackage{algorithm} 
\usepackage{algorithmic}
\usepackage{float}
\usepackage{lipsum}
\usepackage{multicol}
\usepackage{dblfloatfix}    
\usepackage{tabularx}
\usepackage[T1]{fontenc}
\usepackage[utf8]{inputenc}
\usepackage[english]{babel}
\usepackage[english]{translator}
\let\labelindent\relax
\usepackage{enumitem}
\usepackage{adjustbox}
\setlist{leftmargin=3mm}
 \usepackage{multirow}
\usepackage{amsmath}
 \usepackage{lscape}
\PassOptionsToPackage{table}{xcolor}
\usepackage{booktabs, array}
\usepackage{colortbl}
\newcolumntype{Y}{>{\raggedright\arraybackslash}X}

\definecolor{headergray}{gray}{0.85}
\definecolor{rowgray}{gray}{0.95}

\usepackage[nomain,acronym, toc, shortcuts, translate=babel]{glossaries}
\makeglossaries
\loadglsentries[type=\acronymtype]{acro.tex}

\def\BibTeX{{\rm B\kern-.05em{\sc i\kern-.025em b}\kern-.08em
    T\kern-.1667em\lower.7ex\hbox{E}\kern-.125emX}}
    
\begin{document}

\title{EcBot: Data-Driven Energy Consumption Open-Source MATLAB Library for Manipulators 
\thanks{}

}
\author{ Juan Heredia$^{1}$, Christian Schlette$^{1}$ and Mikkel Baun Kjærgaard$^{1}$
\thanks{ $^{1}$ The authors are with the Maersk Mc-Kinney Moller Institute of the University of Southern Denmark}
\thanks{This work was funded by the project ``Industry 4.0 lab facilities for Experimenting with Spatial and Electricity Consumption Data”. }
\thanks{Corresponding email: jehm@mmmi.sdu.dk}}

\maketitle

\begin{abstract}
Existing literature proposes models for estimating the electrical power of manipulators, yet two primary limitations prevail. First, most models are predominantly tested using traditional industrial robots. Second, these models often lack accuracy. To address these issues, we introduce an open source Matlab-based library designed to automatically generate \ac{ec} models for manipulators. The necessary inputs for the library are Denavit-Hartenberg parameters, link masses, and centers of mass. Additionally, our model is data-driven and requires real operational data, including joint positions, velocities, accelerations, electrical power, and corresponding timestamps. We validated our methodology by testing on four lightweight robots sourced from three distinct manufacturers: Universal Robots, Franka Emika, and Kinova. The model underwent testing, and the results demonstrated an RMSE ranging from 1.42 W to 2.80 W for the training dataset and from 1.45 W to 5.25 W for the testing dataset. 
\end{abstract}


\section{Introduction}

In modern society, the industrial sector has experienced a paradigm shift, moving beyond a sole focus on generating high revenues to promoting sustainable production practices \cite{united_nations_17_nodate,seliger_approaches_2008}. This shift coincides with the growing attention given to robots in various industries, \acp{ir}, \acp{cobot} or \acp{lwr}. The integration of robots into industrial settings offers the potential to enhance production rates while concurrently reducing \ac{ec} through environmentally conscious operations and energy-efficient programming \cite{ogbemhe_achieving_2017}.

While industrial robots have been employed in the industry for a considerable period, with the first robot installed nearly 60 years ago \cite{gasparetto_brief_2019}, a new generation of manipulators has emerged, bringing about a notable market transformation. The distinguishing feature of these new robots is their ability to operate alongside humans in the same environment. To enable this collaboration, these robots are specifically designed with lighter materials and reduced stiffness, allowing them to absorb energy during collisions and minimize potential damage \cite{albuschaffer_dlr_2007}. Additionally, they incorporate safety protocols aimed at reducing hazards in shared workspaces. These robots are commonly referred to as collaborative robots or lightweight robots. It is important to note that the terms "collaborative robots" \cite{international_federation_of_robotics_demystifying_2020} and "lightweight robots" \cite{bauer_lightweight_2016} are often used interchangeably in the existing literature.

 \begin{figure}[t]
  \centering
  \includegraphics[width=0.5 \textwidth]{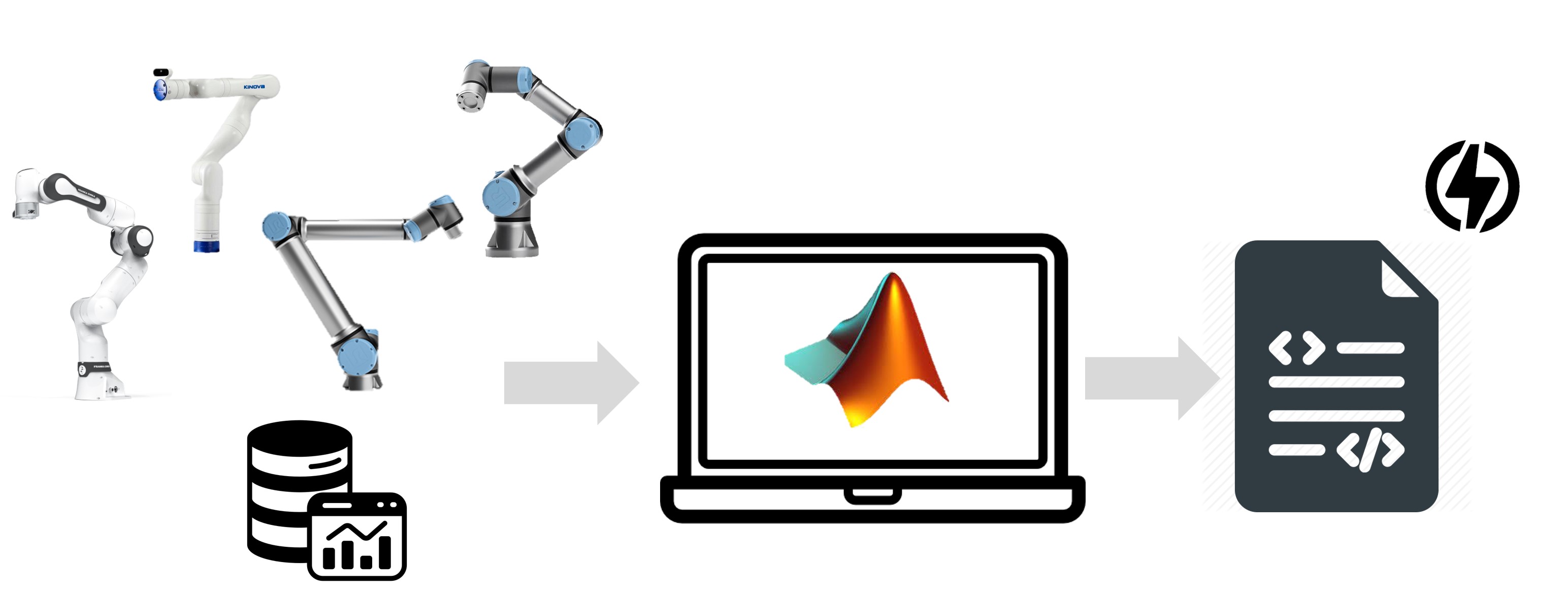}
  \caption{EcBot MATLAB Library for Dynamic and Energy Consumption Modeling for Manipulators.}
  \label{fig:abstract}
  \vspace{-0.6cm}
\end{figure} 

The focus of this study is to address the significant challenges prevalent in the field of \ac{ec} modeling for cobots. The analysis begins with a review of the existing academic literature, which predominantly focuses on \acp{ir}, highlighting a significant research gap concerning cobots. This \ac{ec} modeling gap is particularly concerning considering the findings of Gadaleta et al. (2020) \cite{paper11}, who argue that current \ac{ec} models for \acp{ir} lack reliability. Furthermore, they report that errors in these models are often concealed in academic literature, exacerbating the problem. The complexity of modern robotics, particularly cobots, significantly contributes to these errors. Verstraten et al. (2016) \cite{verstraten_energy_2016} underscore this issue, discussing accuracy problems in \ac{ec} models even for a single DC motor. As the focus shifts to manipulators, these accuracy issues become exponentially more problematic due to the increased complexity of dynamic models, making it challenging to estimate \ac{ec} accurately.


To address these challenges, we previously introduced a modeling approach for estimating the \ac{ec} of cobots in our prior work \cite{heredia_data-driven_2021}. However, this methodology was only validated on two Universal Robots, specifically the UR10e and UR3e, which share similar mechanical designs. Moreover, the model's creation involved a semi-manual modeling and training process. When dealing with a robot of different design or structure, the entire process had to be repeated. The novelty of this article lies in the introduction of an automated library that generates, trains, and tests \ac{ec} models for a wide range of manipulators. This library necessitates specific data inputs, including Denavit-Hartenberg parameters, the mass and center of gravity of the robot's links, and operational data (refer to Fig. \ref{fig:abstract}). This software solution is accessible online at \cite{herediagitlab}, where \ac{ec} models for robots of varying degrees of freedom (DoF) and geometries can be trained and assessed. Additionally, in this paper, we present four case studies where the library is implemented for four distinct robot designs. This highlights the library's versatility and its applicability to various robot architectures and specifications.

The structure of the paper is as follows: Section II provides an overview of the state-of-the-art models from the literature. Section III presents our model in detail. In Section IV, we present case studies showcasing the \ac{ec} model results for five robots from three different manufacturers: UR3e, UR5e, UR10e, FR3, and Gen3. Finally, Section V presents the conclusions of the article.

\section{Related Work}
In the literature, there have been several attempts to model the  \ac{ec} of mechatronic systems, ranging from simple 1 DoF mechanisms \cite{oliva_engineering_2016,verstraten_energy_2016} to more complex manipulators such as \acp{ir} and \acp{cobot}. In this analysis, we will examine the principal attempts from the literature to model the \ac{ec} of these systems:

 \begin{figure*}[h]
  \centering
  \includegraphics[width=0.7 \textwidth]{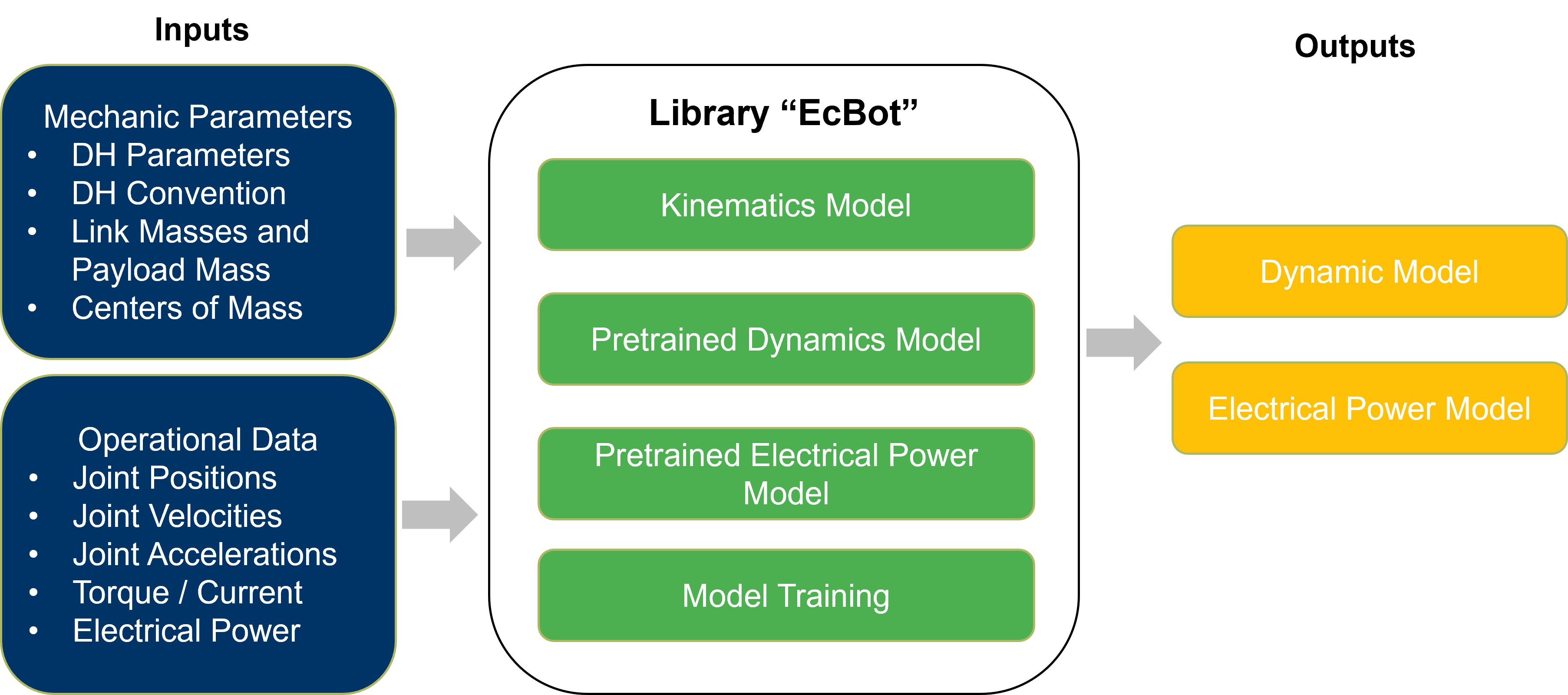}
  \caption{EcBot: Energy Consumption Data-Driven Modeling Library}
  \label{fig:modelinglibrary}
\end{figure*} 

\begin{itemize}
    \item Authors in \cite{paper1} present an \ac{ec} model for the ABB IRB 1200 \ac{ir}. This model focuses on identifying inertia and friction parameters based on simulated data. However, its limitation lies in relying exclusively on simulated data from the manufacturer's simulation tool, "RoboStudio," for training the model. Additionally, the model only considers the \ac{ec} of the robot's motors, neglecting other components.
    \item In a study by \cite{paper3}, an \ac{ec} model for the Motoman MH5L \ac{ir} is developed using the "Catia Systems Engineering" simulation tool. This model provides a mechanical representation of the robot's structure and incorporates both mechanical and electrical losses of each motor. However, the testing conducted for this model was limited to short experiments lasting up to 5 seconds. Longer tests are necessary for comprehensive validation.
    \item Another \ac{ec} model for the KR200 \ac{ir} is formulated in a study by \cite{paper7}. This model encompasses the \ac{ec} of various components, including the manipulator elements, drive system, and auxiliary components like the computer and cooling system. However, the paper lacks an accuracy analysis as the model has not been subjected to any tests.
    \item A specific \ac{ec} model for manipulators, using the ABB IRB140 as a case study is presented by \cite{paper8}. The methodology involves a three-step process: kinematics, dynamics, and \ac{ec} calculation. However, this model only considers the \ac{ec} of the motors, excluding other robot components. Moreover, the validity of the model was not confirmed with real data.
    \item Eggers et al. \cite{eggers_modeling_2018} present an \ac{ec} model that incorporates the temperature of the motors. The study uses the KR16 \ac{ir} to investigate how the robot's operational temperature influences its \ac{ec}. The results demonstrate that the \ac{ec} reduces as the oil temperature of the gearbox increases due to reduced viscosity and mechanical losses. The presented model increases accuracy by accounting for motor temperature.
    \item In a study by \cite{paper15}, an \ac{ec} modeling methodology for manipulators is presented and tested using the ABB 1200 \ac{ir}. The model is then used to detect faults in the robot's operation under different scenarios.
\end{itemize}

The existing literature on \ac{ec} modeling for \acp{ir} reveals several limitations that require further research. Firstly, there is an over-reliance on simulated data for model development and validation, as seen in the study by \cite{paper1}. This practice may limit the models' real-world applicability given the complexity and unpredictability of actual operational environments. Secondly, many studies have focused narrowly on the \ac{ec} of specific components, primarily motors, disregarding other components that significantly contribute to overall energy usage \cite{paper1, paper8}. This selectivity may result in less comprehensive and potentially less accurate models. Thirdly, there is a recurring issue of insufficient model validation. For example, the study in \cite{paper7} did not conduct any testing, and the one in \cite{paper8} was not validated with real data. Without robust validation, the reliability and practical applicability of these models remain questionable. Experimental testing in studies like \cite{paper3} is often short-term, lasting up to only 5 seconds. Longer experiments are necessary to assess the models' performance under different operational conditions and over extended periods. 

In \cite{heredia_data-driven_2021}, the authors present a model specifically targeting collaborative robots, or \acp{cobot}. This model has been trained and tested using 55,000 samples to assess both its validity and reliability. While the methodology can theoretically be extended to any manipulator, it has only been empirically tested on 6 DoF manipulators from Universal Robots. Moreover, the process is manual and requires adjustments to align with the specific DoF and parameters of a different robot design.

\section{Modeling Methodology}
The library creates a data-driven model that estimates the robot's \ac{ec} and is available online in \cite{herediagitlab}. The equations are based on the model presented in \cite{heredia_data-driven_2021}. The library generates a generic mathematical model and then trains the model by estimating the unknown parameters. Algorithm \ref{algo} summarizes all the steps performed by the library to generate and train the dynamic, and electric power model of the robot.

The model parameters are identified for each robot design and manufacturer, meaning that each robot has a unique model. Operational data is essential for fine-tuning the model to the specific characteristics of each robot. Consequently, a model developed for one robot design cannot be applied to others. Moreover, even within the same robot design and manufacturer, two robots from the same series might exhibit slight differences, potentially affecting the model's performance.

\begin{algorithm}
\caption{Generate Model}
\label{algo}
\begin{algorithmic}
\STATE Initialize gravity $g = 9.8$, payload mass $m_p$, payload momentum $n_{l+1}$
\STATE Define number of joints $N_{art}$
\STATE Define the DH Convention

\FOR {$i = 1$ to $N_{art}$}
    \STATE Generate matrices $^{i-1}T_i$ and $^{i-1}R_i$
    \STATE Calculate kinematic vectors $r_{i,i+1}$, $r_{i+1,c}$, $w_i$, $\dot{w}_i$, $\ddot{p}_i$ and $\ddot{p}_{Ci}$ 
    \STATE Update transformation matrix $^{0}T_i = ^{0}T_{i-1} \times  ^{i-1}T_{i}$
\ENDFOR

\FOR {$i = N_{art}$ to $i = 1$}
    \STATE Calculate link force $f_i$
    \STATE Calculate known momentum $n_{ki}$
    \STATE Calculate unknown momentum $n_{ui}$
    \STATE Calculate torque ($\tau_{i}$) or current  $(i_{i})$
\ENDFOR

\FOR {$i = 1$ to $N_{art}$}
    \STATE Train the torque/current model ($M_i = \Theta_i K_i^T$) by estimating the unknown parameters ($\Theta_i$) based on the known operational data ($K_i$) :
    \STATE $K_i = (\Theta_{i}^T\Theta_{i})^{-1}\Theta_{i}^T{M_{i}}$
\ENDFOR   
\STATE Generate the electrical power model ($P_{R} = K_{P} \Theta_{P}^T$)
\STATE Train the model by estimating the unknown parameters:
\STATE $K_{P} = (\Theta_{P}^T\Theta_{P})^{-1}\Theta_{P}^T \, P_{Rm}$
\STATE Return the dynamic and electrical model
\end{algorithmic}
\end{algorithm}

\subsection{Requirements}
The model requires the following information:

\begin{itemize}
    \item Denavit–Hartenberg parameters
    \item Denavit–Hartenberg convention
    \item Link masses
    \item Link centers of mass
    \item Payload Mass
    \item Operational Data: Articulation position, velocity, and acceleration of each joint, joint torque/motor current, electrical power, and the time stamp.     
\end{itemize}

\subsection{Mathematical Model}

\subsubsection{Kinematics Model} The robot kinematics of the manipulator is defined according to the Denavit–Hartenberg parameters. Two conventions use two different homogenous matrices. The library can use any of the conventions. 

\textit{Traditional Convention:} This convention transform the system $O_{i-1}$ to $O_{i}$ using the following four movements:

\begin{itemize}
    \item $d_n$ : offset along $z_{i-1}$ ,
    \item $\theta_n$ : angular offset about $z_{i-1}$ to align $x_{i-1}$ and $x_{i}$ ,
    \item $a_n$ : offset along $x_{i-1}$ ,
    \item $\alpha_n$ : angular offset along $x_{i-1}$ to align $z_{i-1}$ and $z_{i}$ ,
\end{itemize}

such that :

\vspace{-1mm}
\begin{equation}
\resizebox{0.4\textwidth}{!}{
$^{n-1}T_n= \begin{bmatrix}
\cos\theta_n & -\sin\theta_n \cdot \cos\alpha_n & \sin\theta_n \cdot \sin\alpha_n &       a_n \cdot \cos\theta_n \\
 \sin\theta_n   &  \cos\theta_n \cdot \cos\alpha_n   &   -\cos\theta_n\cdot\sin\alpha_n   &    a_n \cdot  \sin\theta_n \\
 0            &          \sin\alpha_n    &              \cos\alpha_n    &               d_n\\
 0            &           0            &    0        &    1
\end{bmatrix} $ .
}
\end{equation}

 \vspace{-1mm}

\textit{Modified Convention:} This convention transform the system $O_{i-1}$ to $O_{i}$ using the following four movements:

\begin{itemize}
    \item $\alpha_n$ : angular offset along $x_{i-1}$ to align $z_{i-1}$ and $z_{i}$,
    \item $a_n$ : offset along $x_{i-1}$,
    \item $\theta_n$ : angular offset about $z_{i-1}$ to alling $x_{i-1}$ and $x_{i}$ ,
    \item $d_n$ : offset along $z_{i-1}$,
\end{itemize}

such that :
\vspace{-1mm}
\begin{equation}
\resizebox{0.4\textwidth}{!}{
$^{n-1}T_n= \begin{bmatrix}
\cos\theta_n & -\sin\theta_n  & 0 &       a_{n-1}  \\
\sin\theta_n \cdot \cos\alpha_{n-1}  &  \cos\theta_n \cdot \cos\alpha_{n-1} & - \sin\alpha_{n-1}   &   - d_n \cdot \sin\alpha_{n-1}    \\
\sin\theta_n \cdot \sin\alpha_{n-1}  &  \cos\theta_n \cdot \sin\alpha_{n-1} & \cos\alpha_{n-1}   &   - d_n \cdot \cos\alpha_{n-1}  \\
 0            &           0            &    0        &    1
\end{bmatrix}$ } .
\end{equation}
\vspace{-1mm}

\subsubsection{Dynamics Model} 
The robot's dynamics model is established using the Newton-Euler formulation, as detailed in the work by \cite{rigelsford_modelling_2000}. This formulation involves an equilibrium analysis of all the forces acting on each link to determine the joint torques required for each articulation. 

\textbf{Forward Recursion}: Initially, the method performs a forward recursion from the first articulation to the l-th articulation, which enables the determination of key parameters for each link, including angular velocity ($w_i$), angular acceleration ($\dot{w}_i$), linear acceleration ($\Ddot{p}_i$), and center of mass acceleration ($\Ddot{p}_{Ci}$). The joint velocity is determined as follows:

\vspace{-4mm}
\begin{equation}
    w_i=w_{i-1} + \dot{\theta}_i u_{i}  \, ,
\end{equation}

where $u_{i}$ represents the $i$ - joint rotation unit vector, and it is determined by $u_i = R_i \, z_0$. The vector $z_0$ denotes the orientation of the z-axis within the global coordinate system, and it is defined as $z_0 = (0, 0, 1)$. Subsequently, the remaining kinematic vectors are calculated through the application of the following equations:

\begin{equation}
    \dot{w}_i= \dot{w}_{i-1} + \Ddot{\theta_i} z_0 + \dot{\theta}_i w_{i-1} \times u_{i} \, ,
\end{equation}
\begin{equation}
    \Ddot{p}_i= \Ddot{p}_{i-1} + \dot{w}_i \times r_{i,i+1} + w_i \times ( w_i \times r_{i,i+1}) \, ,
\end{equation}
\begin{equation}
    \Ddot{p}_{C_i}=  \Ddot{p}_i + \dot{w}_i \times r_{i+1,C_i} + w_i \times ( w_i \times r_{i+1,C_i}) \, .
\end{equation}

The initial conditions are set as follows: angular velocity at the base ($w_0$) is [0, 0, 0], angular acceleration at the base ($\dot{w_0}$) is [0, 0, 0], linear acceleration at the base ($\Ddot{p}_0$) is [0, 0, 0], and center of mass acceleration at the base ($\Ddot{p}_{C0}$) is [0, g, 0].



\textbf{Backward Recursion}: This approach proceeds with a reverse recursion, facilitating the transmission of forces and moments from \textit{l-th} articulation back to the first articulation. The force exerted on each link, denoted as $f_i$, is determined based on factors including the preceding forces $f_{i+1}$, the mass of the respective link, denoted as $m_i$, and the linear acceleration of the center of mass, characterized as $\Ddot{p}_{Ci}$, such that:

\vspace{-3mm}
\begin{equation}
    f_i = f_{i+1} + m_i \Ddot{p}_{C_i} \, .
\end{equation}

Then, the moment vectors, referred to as $n_i$, are calculated. The initial conditions are specified as $f_{l+1} = [fx, fy, fz]$ and $n_{l+1} = [mx, my, mz]$. Here, the variables $fx, fy, fz, mx, my, mz$ correspond to the forces and moments applied to the robot's Tool Center Point (TCP) by the payload.
 \vspace{-2mm}
\begin{equation}
\begin{split}
    n_{i} = & n_{i+1} + r_{i-1,i} \times f_{i+1}\\
   &  + r_{i,C_i} \times m_i \Ddot{p}_{C_i} + I_i \dot{w}_i + w_i \times I_i w_i  \, ,
\end{split}
\end{equation}
where $I_i$ is the i-link inertia. 



In our mathematical model, we consider a momentum vector, represented as $n_{i+1}$. This vector pertains to a previous articulation and can be subdivided into two primary components. The first component comprises momentum characterized by known parameters, denoted as $n_{k, i}$, while the second encompasses momentum characterized by unknown parameters, symbolized as $n_{u, i}$. As a result, the momentum vector $n_{i+1}$ can be succinctly expressed as the sum of these known and unknown parameters, that is, $n_{i+1} = n_{k, i+1} + n_{u, i+1}$.

Moreover, the model comprises another momentum vector, specifically associated with the tool and denoted by $n_{l+1}$. This vector is considered null, signifying that we perceive the tool as a static entity. Then, the momentum vector is segregated and formulated as:

\vspace{-4mm}
\begin{equation}
     n_{k \, i} = \sum_{j=i}^l ( r_{j-1,j} \times f_{j+1} + r_{j,C_j} \times m_j \Ddot{p}_{C_j} )\, , 
\end{equation}

\vspace{-4mm}
\begin{equation}
     n_{u \, i} = \sum_{j=i}^l (I_j \dot{w}_j + w_j \times I_j w_j) \, .  
\end{equation}

The determination of motor torques involves projecting the moment $n_i$ onto the rotation axis $u_i$. Additionally, we include the joint viscous torque ($ F_{vi} = k_{vi} \dot{\theta}i$) and the joint Coulomb torque ($F{si} = k_{si} \text{sgn}(\dot{\theta}_i)$), where the parameters $ k_{vi} $ and $k_{si}$ represent joint viscous and Coulomb friction constants. These friction elements are incorporated into the following equation:

\vspace{-3mm}
\begin{equation}
    \tau_i = n_{k \, , i} \cdot u_{i} + n_{u \, , i} \cdot u_{i}  + k_{vi} \dot{\theta}_i + k_{si}sgn(\dot{\theta}_i) \, .
\end{equation}


Then, the subsequent step involves the calculation of the corresponding motor current, symbolized by $i$. This is facilitated by the relationship between torque and current, represented as $\tau_m = k_m \cdot i$. The resultant equation is:

\vspace{-2mm}
\begin{equation}
    i_{i} = \frac{1}{k_m} ( n_{k \, , i} \cdot u_{i} + n_{u \, , i} \cdot u_{i}  + k_{vi} \dot{\theta}_i + k_{si}sgn(\dot{\theta}_i)) \, .
    \label{eq.curr}
\end{equation}

\subsubsection{Power Model} The total power consumption is determined by summing up the power usage of all individual components, namely, the electronics components' power($P_{E}$), mechanical brakes' power ($P_{B}$), braking resistance power ($P_{R}$), each motor power ($P_{motor_i}$), and motor driver power ($P_{MD_i}$) \cite{heredia_ecdp_2023}. As a result, we propose the following equation to calculate the electrical power:

\vspace{-4mm}
\begin{equation}
    P_{R} \,= \,P_{E} + P_{B} + P_{R} +  \sum_{i = 1}^{l} (P_{motor_i} + P_{MD_i}) \, .
\label{eq:Eq1}
\end{equation}


Each power component is mathematically modelled, and they are described as follows:
\begin{itemize}

    \item Electronic Components' Consumption $P_{E}$: This consumption component is complex, and its \ac{ec} fluctuates less than that other components such a motor. Thus, the power consumption of the electronic devices is treated as a constant, denoted as $P_{E} = cte$.
    
    \item Mechanical Brakes' Consumption $P_B$: The brakes remain continuously engaged to enable the movement of the robot. Consequently, the power consumption of the brakes is also considered constant and is expressed as $P_{B} = cte$. This leads to the combined variable $P_{c}$, which is the sum of $P_{E}$ and $P_{B}$ and remains constant throughout robot operation.

    \item Driver Consumption: The motor driver consumption is directly proportional to the absolute value of the motor current , as given by the equation: 

    \begin{equation}
        P_{MD} = k_{MD} \cdot |i| \, ,
    \end{equation}

    where $P_{MD}$ represents the power consumption of the motor driver, and $k_{MD}$ is the proportionality constant associated with the motor driver.

    \item Motors' Consumption $P_{motor}$ : Motor types vary between different manufacturers. Consequently, we employed a universal representation of the motor, conceptualizing it as a combination of inductance, resistance, and a voltage source. The electrical equations encapsulating the motor model are as follows:

    \begin{equation}
    \begin{cases}
        P_{motor} = i (L \frac{di}{dt} + R i + e)\\
        P_{motor} = \frac{1}{k_m} \tau_m (\frac{L}{k_m} \frac{d \tau_m}{dt} + \frac{R}{k_m} \tau_m + e)
    \end{cases}
    \end{equation}
    
    where, $i$, $e$, $\tau_m$, and $k_m$ are the motor current, induced voltage, motor torque, and motor torque constant respectively. 

    \item Regenerative Braking Consumption ($P_R$) is in charge of restricting the electrical power of the robot, ensuring that the cobot's power consumption consistently remains positive. The robot's energy system is not designed to restore energy to the system, nor does it have a battery to store this energy.  
    
\end{itemize}

By substituting the mathematical models of each power component, we derive two equations. The decision on whether to use Equation \ref{eq.power} or Equation \ref{eq.power2} to estimate the robot's electrical power depends on the sensor configuration within the robot. To be more specific, if the robot incorporates current sensors, Equation \ref{eq.power} is the suitable option, whereas if torque sensors are present in the robot, it is advisable to employ Equation \ref{eq.power2}.

\begin{equation}
\begin{split}
    P_{R}  = & P_{c} + 
    \sum_{i = 1}^{l} ( L_{i} i_{i} \frac{di_{i}}{dt} + 
    R_{i}  i_{i}^2   + k_t  \dot{\theta}_i   i_{i} + k_{MDi} |i_{i}|) 
\label{eq.power}
\end{split}
\end{equation}

\begin{equation}
   P_{R}  = P_{c} + 
    \sum_{i = 1}^{l} ( \frac{L_{i}}{k_{mi}^2}  \tau_{i} \frac{d\tau_{i}}{dt} + 
    \frac{R_{i}}{k_{mi}^2}  \tau_{i}^2   +  \frac{ k_t}{k_{mi}} \dot{\theta}_i   \tau_{i} +  \frac{ k_{MDi}}{k_{mi}}  |\tau_{i}|) 
\label{eq.power2}
\end{equation}

\subsection{Training}

\textbf{Dynamic Model Training}: The nonlinear equations $i_i$ or $\tau_i$ require a conversion into a multilinear polynomial format. This transformation results in an equation expressed as a linear combination of terms: $i_i = X_0 + \beta_1 X_1 + \beta_2 X_2 + ... + \beta_n X_n$, wherein every $\beta_i$ represents a nonlinear equation including unknown parameters, and each $X_i$ represents a nonlinear equation with known parameters, such that:


 \begin{equation}
 \begin{cases}
    i_{i} = \Theta_i K_i^T \\
    \tau_{i} = \Theta_i K_i^T
    \label{eq:current} 
\end{cases}
\end{equation}
 
where $\Theta_i$ is a vector $\Theta_i= [X_0, X_1, ... X_n]$) dependent on known parameters, expressed as $ X_m = f(\theta, \dot{\theta}, \ddot{\theta}, l, m, r_c)$. Meanwhile, $K_i$ is a vector ($K_i = [1 , \beta_1, ..., \beta_n]$) that is function of unknown  parameters, indicated as $\beta_m = f(I,k_{vi},k_{si},f_{l+1},n_{l+1})$. These two vectors are determined using a recursive method that disaggregates the motor torque or current dynamic model into known and unknown parameters, and presents them in a multilinear form. This process is accomplished with the assistance of the symbolic representation library in Matlab.



Subsequently, the identification of the unknown parameter vector $K_i$ is accomplished through the application of the least mean square identification method, as follows:


 \vspace{-1mm}
 \begin{equation}
    K_i = (\Theta_{i}^T\Theta_{i})^{-1}\Theta_{i}^T{M_{i}}
    \label{eq.iddy}
\end{equation}

where , $\Theta_i$ represents the measurement matrix of the i-th joint, with dimensions $m_{\Theta} \times n_{\Theta}$. Here, $m_{\Theta}$ corresponds to the count of parametric variables within $\Theta_i$, and $n_{\Theta}$ pertains to the number of measurements. Additionally, $M_i$ serves as the measurement matrix for the current or torque of the i-th joint, with dimensions $n_{\Theta} \times 1$. This iterative process is repeated l times, where l signifies the number of DoF.


\textbf{Electrical Power Model}: Similarly to how the dynamic model operates, the unknown parameters in the electrical power model are determined. The robot's power, labeled as $P_{R}$, is expressed as a linear combination of terms that are either known or unknown. The grouping of these parameters is achieved using the following equation:


\begin{equation}
P_{R} = K_{P} \Theta_{P}^T \, ,
\label{eq:power}
\end{equation}

where ${K}_{P} = [P{c}, L_{1}, R_{1}, k_{t_1}, k_{MD1}, ... , L_{l}, R_{l}, k_{t_l}, k_{MDl}]$ and ${\Theta}{P} = [1, i{1}\frac{di_{1}}{dt} , i_{1}^2, \dot{\theta}1|i{1}|, ... , i_{l}\frac{di_{l}}{dt} , i_{l}^2, \dot{\theta_l}|i{l}|]$. 
For example, in the case of a 7 DoF manipulator, the vectors ${K}{P}$ and ${\Theta}{P}$ each have dimensions of (29 x 1).

The determination of not known parameters was achieved through the application of the least mean square identification method, resulting in the derivation of the parameter vector $K_{P}$, which is obtained using the following equation:


\begin{equation}
K_{P} = (\Theta_{P}^T\Theta_{P})^{-1}\Theta_{P}^T \, P_{Rm} \, ,
\label{eq.powerpara}
\end{equation}

where $\Theta_{Pm}$ represents the measurement matrix of $\Theta_{P}$. It has dimensions of $n_{\Theta_{Pm}} \times m_{\Theta_{Pm}}$, where $n_{\Theta_{Pm}}$ signifies the count of parametric variables in $\Theta_{P}$ (which is 29 for a 7 Degrees of Freedom robot), and $m_{\Theta_{Pm}}$ represents the number of measurements. Additionally, $P_{Rm}$ is the matrix for power measurements, and it has dimensions of $m_{\Theta_{Pm}} \times 1$.


\begin{table*}[b]
\vspace{-2mm}
\centering
\caption{Training parameters for robots UR3e, UR10e, Gen3, and FR3.}
\label{Table_1}
\vspace{-2mm}
\setlength{\tabcolsep}{6pt}
\renewcommand{\arraystretch}{1.3}
\resizebox{0.94\textwidth}{!}{
\begin{tabularx}{1.5\textwidth}{X X X X}
\rowcolor{headergray}
\textbf{UR3e} & \textbf{UR10e} & \textbf{Gen3} & \textbf{FR3} \\

$\begin{array}{@{}l@{}}
d = [0.151, 0, 0, 0.131, 0.085, 0.092]\,\mathrm{m} \\
a = [0, -0.244, -0.213, 0, 0, 0]\,\mathrm{m} \\
\alpha = [\pi/2, 0, 0, \pi/2, -\pi/2, 0]\,\mathrm{rad} \\
m = [1.98, 3.44, 1.44, 0.87, 0.81, 0.26]\,\mathrm{kg} \\
m_{\mathrm{payload}} = 0\,\mathrm{kg} \\
DH = 0 \\
r_{c1} = [0,-0.02,0]\,\mathrm{m} \\
r_{c2} = [0.13,0,0.12]\,\mathrm{m} \\
r_{c3} = [0.05,0,0.02]\,\mathrm{m} \\
r_{c4} = [0,0,0.01]\,\mathrm{m} \\
r_{c5} = [0,0,0.01]\,\mathrm{m} \\
r_{c6} = [0,0,-0.02]\,\mathrm{m}\\
\\
\\
\\
\\
\end{array}$
&

$\begin{array}{@{}l@{}}
d = [0.181, 0, 0, 0.174, 0.120, 0.112]\,\mathrm{m} \\
a = [0, -0.613, -0.572, 0, 0, 0]\,\mathrm{m} \\
\alpha = [\pi/2, 0, 0, \pi/2, -\pi/2, 0]\,\mathrm{rad} \\
m = [7.37, 13.05, 3.99, 2.10, 1.98, 0.62]\,\mathrm{kg} \\
m_{\mathrm{payload}} = 0\,\mathrm{kg} \\
DH = 0 \\
r_{c1} = [0.021,0,0.27]\,\mathrm{m} \\
r_{c2} = [0.38,0,0.16]\,\mathrm{m} \\
r_{c3} = [0.24,0,0.07]\,\mathrm{m} \\
r_{c4} = [0,0.01,0.02]\,\mathrm{m} \\
r_{c5} = [0,0.01,0.02]\,\mathrm{m} \\
r_{c6} = [0,0,-0.03]\,\mathrm{m} \\
\\
\\
\\
\\
\end{array}$
&

$\begin{array}{@{}l@{}}
d = [0, -0.18, -0.01, -0.42, 0.02, -0.31,  \\
\hspace{5mm}     0, -0.17]\,\mathrm{m} \\
a = [0,0,0,0,0,0,0,0]\,\mathrm{m} \\
\alpha = [\pi, \pi/2, \pi/2, \pi/2, \pi/2, \pi/2, \pi/2, \pi]\,\mathrm{rad} \\
m = [1.70, 1.38, 1.16, 1.16, 0.93, 0.68, 0.68,\\
\hspace{7mm} 0.50]\,\mathrm{kg} \\
m_{\mathrm{payload}} = 0\,\mathrm{kg} \\
DH = 0 \\
r_{c1} = [0,0,0.08]\,\mathrm{m} \\
r_{c2} = [0,-0.01,0.21]\,\mathrm{m} \\
r_{c3} = [0,0.03,-0.01]\,\mathrm{m} \\
r_{c4} = [0,0,0.30]\,\mathrm{m} \\
r_{c5} = [0,0.13,-0.02]\,\mathrm{m} \\
r_{c6} = [0,-0.01,0.25]\,\mathrm{m} \\
r_{c7} = [0,0.10,-0.02]\,\mathrm{m} \\
r_{c8} = [0,-0.01,0.04]\,\mathrm{m}
\end{array}$
&

$\begin{array}{@{}l@{}}
d = [0.333, 0, 0.316, 0, 0.384, 0, 0.107]\,\mathrm{m} \\
a = [0,0,0,0.0825,-0.0825,0,0.088]\,\mathrm{m} \\
\alpha = [0,-\pi/2,\pi/2,\pi/2,-\pi/2,\pi/2,\pi/2]\,\mathrm{rad} \\
m = [4.97,0.65,3.23,3.59,1.23,1.67,0.74]\,\mathrm{kg} \\
m_{\mathrm{payload}} = 0\,\mathrm{kg} \\
DH = 1 \\
r_{c1} = [0.003,0.002,0]\,\mathrm{m} \\
r_{c2} = [-0.003,-0.02,0.003]\,\mathrm{m} \\
r_{c3} = [0.02,0.04,-0.06]\,\mathrm{m} \\
r_{c4} = [-0.05,0.10,0.02]\,\mathrm{m} \\
r_{c5} = [-0.01,0.04,-0.04]\,\mathrm{m} \\
r_{c6} = [0.06,-0.01,-0.01]\,\mathrm{m} \\
r_{c7} = [0.01,-0.04,0.06]\,\mathrm{m}\\
\\
\\
\\

\end{array}$ 
\\
\end{tabularx}
}
\end{table*}

\subsection{Library Functions}
\subsubsection{Model Generation and Training} This function is responsible for generating a symbolic representation of the robot's \ac{ec}. It disaggregates the non-linear model into a multilinear polynomial equation, identifies the unknown parameters, and finetunes the model to match the robot's specific characteristics based on operational data. The mathematical foundation and modeling process are explained in the above sections (Section III B and Section III C).

This data-driven function relies on real-world data collected during the robot's operation, including joint positions, velocities, accelerations, current or torque readings, and the robot's \ac{ec}. By leveraging this comprehensive dataset, the function ensures that the model accurately represents the robot's unique operational behavior, making it a valuable tool for enhancing control and optimization strategies.

The model training process is initiated through the execution of the following command: 

\scalebox{0.7}{
$GenTrain\_Model(d,a,alfa,DH, m, r_c, m_{load}, q,dq,ddq,I,PC,t,name)$.
}

This command takes a range of essential input parameters into account, such as DH parameter distances 'd' and 'a', the DH parameter angle '$\alpha$', the choice of DH parameter convention (0 for normal, 1 for modified), a vector describing link masses 'm', the center of mass for each link denoted as '$r_c$', the mass of the payload '$m_{load}$', joint positions '$\theta$', joint velocities '$\dot{\theta}$', joint accelerations '$\Ddot{\theta}$', joint currents or torques 'I', power consumption 'PC', the timestamp 't', and the designation of the model by its 'name'. These parameters allow for the fine-tuning of the model to the specific characteristics of the robot, ensuring the accuracy and effectiveness of the training process.





\subsubsection{Testing Model}: This function presupposes the existence of a previously trained model. Leveraging the trained model in conjunction with additional datasets, this function evaluates the model's performance. It relies on essential data inputs, including joint positions, velocities, accelerations, timestamps, and the robot's \ac{ec}. In return, the function provides comprehensive accuracy and error metrics that effectively quantify the model's performance, offering valuable insights into its effectiveness and reliability.

This command serves to evaluate the model using a test dataset. It yields the electrical power, the joint current, or torque, along with three evaluation metrics in the vector $e_{P}$. These metrics evaluate the power estimation and comprise the Root Mean Standard Error (RMSE) in Watts [W] (Equation \ref{RMSE}), the relative error percentage RMSE\% (Equation \ref{RMSEa}), and the coefficient of determination, also known as the R-squared value.

\begin{equation}
    RMSE(X) = \sqrt{ \frac{\sum^n(X_{real}-X_{est})^2}{n}}
\label{RMSE}
\end{equation}

\begin{equation}
    RMSE\%(X) = \frac{RMSE(X)}{max(X_{real})-min(X_{real})} 100 \%
\label{RMSEa}
\end{equation}

The command itself is structured as follows:

\scalebox{0.8}{
$[PE, I, e_{P}] = Test\_Model(name,q,dq,ddq,t,PC)$.
}

The input parameters encompass the model's name, joint positions, joint velocities, joint accelerations, time, and electrical power measurements. The output variables produced by this command consist of Electrical Power Estimation (EP), Current or Torque Estimations (I), and the Error of Power Consumption ($e_{P}$). This evaluation process is crucial for verifying the model's performance and reliability when applied to real-world data.



\subsubsection{Individual Model Estimation} This function is used to test individual points of \ac{ec}. The user introduces the joint positions, velocities, and accelerations, and then obtains the torque/current value and its electrical power. The command is structured as follows:

\scalebox{0.8}{
$[EP, I] = PC\_Model(name,q,dq,ddq).$
}

The input parameters are the model's name, joint positions, joint velocities, joint accelerations. The output variables produced by this command consist of Electrical Power Estimation (EP) on the given time stamp, Current or Torque Estimations (I).



\section{Study Cases}

\begin{figure}[h]
  \centering
  \includegraphics[width=0.48\textwidth]{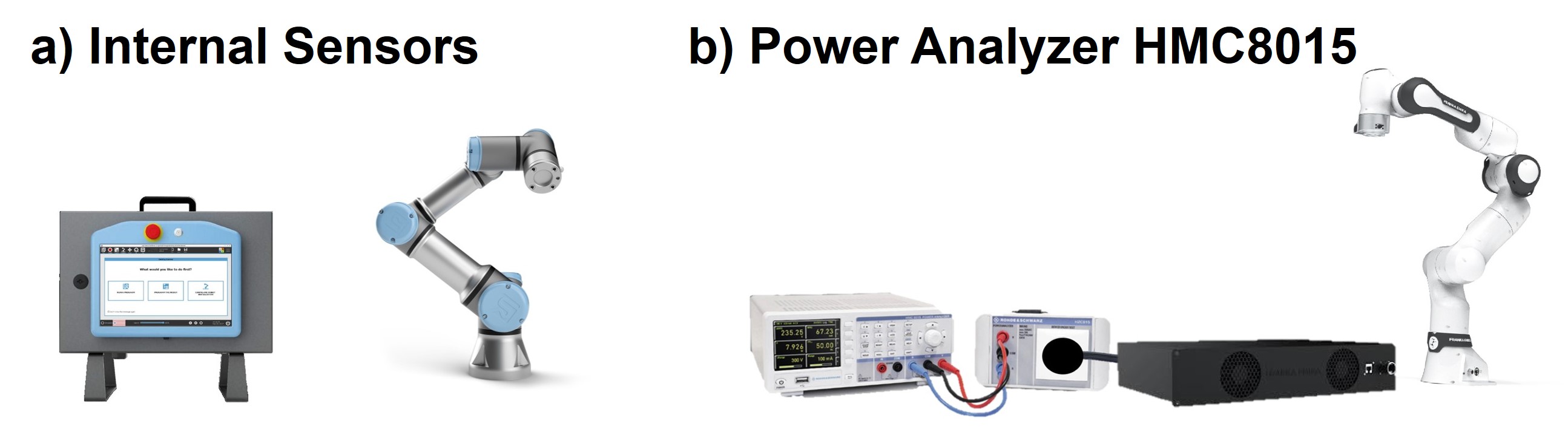}
  \caption{Sensing procedures to measure robot electrical power.}
  \label{fig:sensors}
\end{figure}

In the presented case studies, the efficacy of the model is demonstrated across a variety of robots, including UR3e, UR10e, FR3, and Gen3. Furthermore, we assessed the applicability of the library by employing two separate sensing technologies, as illustrated in Fig. \ref{fig:sensors}. The first approach leverages internal sensors that are manufacturer-provided. Contrarily, in scenarios where robots are devoid of integrated power sensors, an external power analyzer is utilized to measure their electrical power.

The library is tested using four different robots. The parameters of the robots are obtained using the datasheets of the manufacturers \cite{ur_universal_2022,franka_product_2022,kinova_user_2022}. 

 \begin{figure*}[b]
 \vspace{-3mm}
  \centering
  \includegraphics[width=0.78 \textwidth]{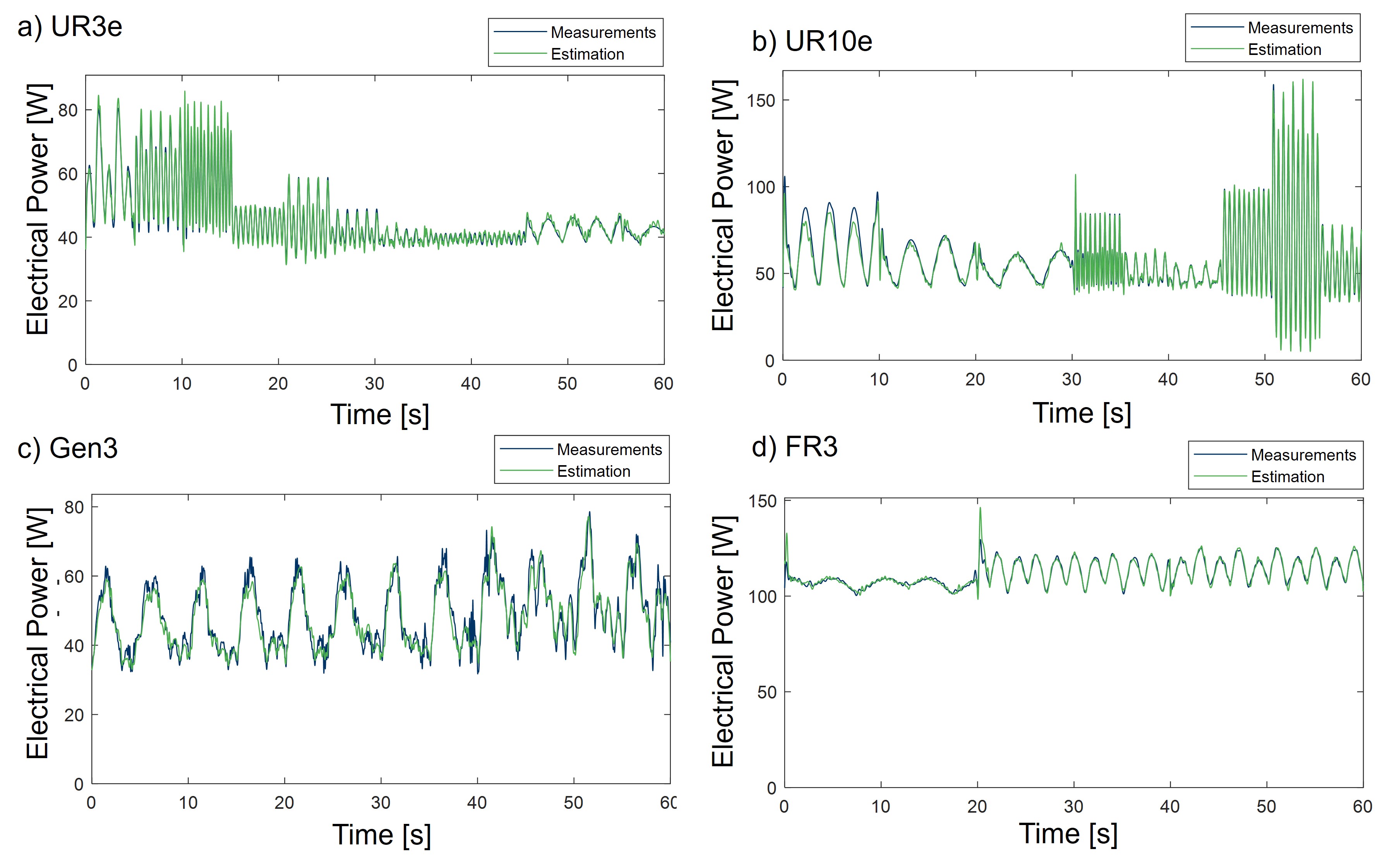}
  \vspace{-4mm}
  \caption{Modeling testing results for four robots, namely, UR3e, UR10e, FR3, and Gen3.}
  \label{fig:results}
\end{figure*}

\begin{LaTeXdescription}
    \item[UR10e and UR3e] Two of the models from Universal Robots were used to train and test the model. Both robots UR3e and UR10e are 6 DoF manipulators manufactured by  Universal Robots, and their maximum payload capacity is 3 kg and 12.5 kg, respectively. The robot parameters required to generate the model are presented in Table \ref{Table_1}. 
    In the case of this manufacturer, the robots have a current and voltage sensor included in their hardware. However, this sensor only measures the \ac{ec} of the manipulator excluding the consumption of the controller or teaching pendant.
    \item[Gen3] The Kinova Gen3 is an ultra-lightweight robot featuring 7 DoF, with a weight-to-payload ratio of 2:1. This robot boasts a payload capacity of 4 kg. According to the manufacturer's design, based on Denavit-Hartenberg parameters, the robot consists of 8 links, as presented in Table \ref{Table_1}. However, the first joint of the design is static. 
    The power analyzer HMC8015 was utilized to measure the robot \ac{ec}.
    \item[ FR3] Franka FR3 is a robot of 7 DoF with a payload capacity of 3 kg. According to the manufacturer's design, the Denavit-Hartenberg parameters are presented in Table \ref{Table_1} using the modified convention.  The power analyzer HMC8015 was utilized to measure the robot \ac{ec}. 
\end{LaTeXdescription}

\subsection{Datasets}

The models underwent testing and training using an extensive dataset of operational data obtained from each of these robots. This data was collected by executing sinusoidal trajectories in each of the robot's joints. The choice of sinusoidal trajectories was deliberate, as they offer versatility and randomness in Cartesian space and guarantee that each joint moves during the experiments. This approach enables us to extract more information from the robot's characteristics and behavior. The joint position follows the next equation:


\vspace{-2mm}
\begin{equation}
    \theta_i = \theta_{i0} + A_i \cdot \sin( 2 \, \pi f_i \, t + \phi_0),
\end{equation}

where $\theta_i$, $\theta_{i0}$, $A_i$, $f_i$, $\phi_0$ are the angular position, angular initial position, oscillation amplitude, oscillation frequency, and oscillation phase of the i-articulation. These values were modified to obtain a diverse dataset that  captures the robot's operational behavior.

The training dataset for each of the robots was recorded obtaining unless 50 000 samples for each robot. In the case of the testing dataset, the robot operation was recorded for additional 60 s. Both datasets are available in \cite{herediagitlab}.

\subsection{Results}

Equation \ref{eq_dyn} is utilized for testing the dynamic model of the library. This process involves the calculation of the average RMSE for all joint torque/current ($I/\tau$) values, as well as the percentage error, denoted as $\% RMSE_D$. Specifically:

\begin{equation}
\small
    RMSE_D  = \sqrt{\sum_{i=1}^{n_{DoF}}\sum^{N} \frac{((I/\tau)_{i, real}-(I/\tau)_{i, est})^2)}{n_{DoF} \, N}},
    \label{eq_dyn}
\end{equation}

\begin{equation}
\small
    \% RMSE_D  =\sum_{i=1}^{n_{DoF}} \frac{RMSE\%((I/\tau)_i)}{n_{DoF}},
    \label{eq_dynp}
\end{equation}

where $n_{DoF}$, $N$ are the number of DoF and the number of measurements,  respectively. The parameters $(I/\tau)_{i, real}$ $(I/\tau)_{i, est}$ are the torque or current of the i-joint for the real and the estimation values. Additionally, RMSE\% are the percentage RMSE determined by Equation \ref{RMSEa}.

   \vspace{-3mm}
\begin{table}[h]

\caption{ Results of the model for the four testing robots}
\label{Tb:results}
\raggedright
\scalebox{0.8}{
\begin{tabular}{ccccccc}
\hline
                                          &          & $RMSE_D$ & $\%RMSE_D$ & RMSE [W] & RMSE\% & $r^2$  \\\hline
\multirow{2}{*}{UR3e}                     & Training & 0.080 A & 2.30 &  1.42    &     1.24   &   0.975    \\
                                          & Testing  & 0.085 A & 3.41 & 1.45    &    2.66    &   0.975   \\\hline
\multirow{2}{*}{UR10e}                    & Training & 0.268 A & 1.71 &2.74    &  1.74      &   0.980    \\
                                          & Testing  & 0.272 A &  2.52 &4.58    &  2.91       &  0.972     \\\hline
\multirow{2}{*}{Gen3}                     & Training & 0.147 N $\cdot$ m & 1.63 & 2.80    & 2.80        & 0.932      \\
                                          & Testing  & 0.275 N $\cdot$ m &  3.07 & 5.25    & 6.55     &  0.894     \\\hline
\multirow{2}{*}{FR3}                      & Training & 0.308 N $\cdot$ m & 5.47 & 2.62   &  3.03      & 0.952      \\
                                          & Testing  & 0.429 N $\cdot$ m & 9.00  & 5.07   &  5.87      &  0.870    \\\hline
\end{tabular}%
}
\end{table}

\begin{figure}[h]
  \centering
  \includegraphics[width=0.4 \textwidth]{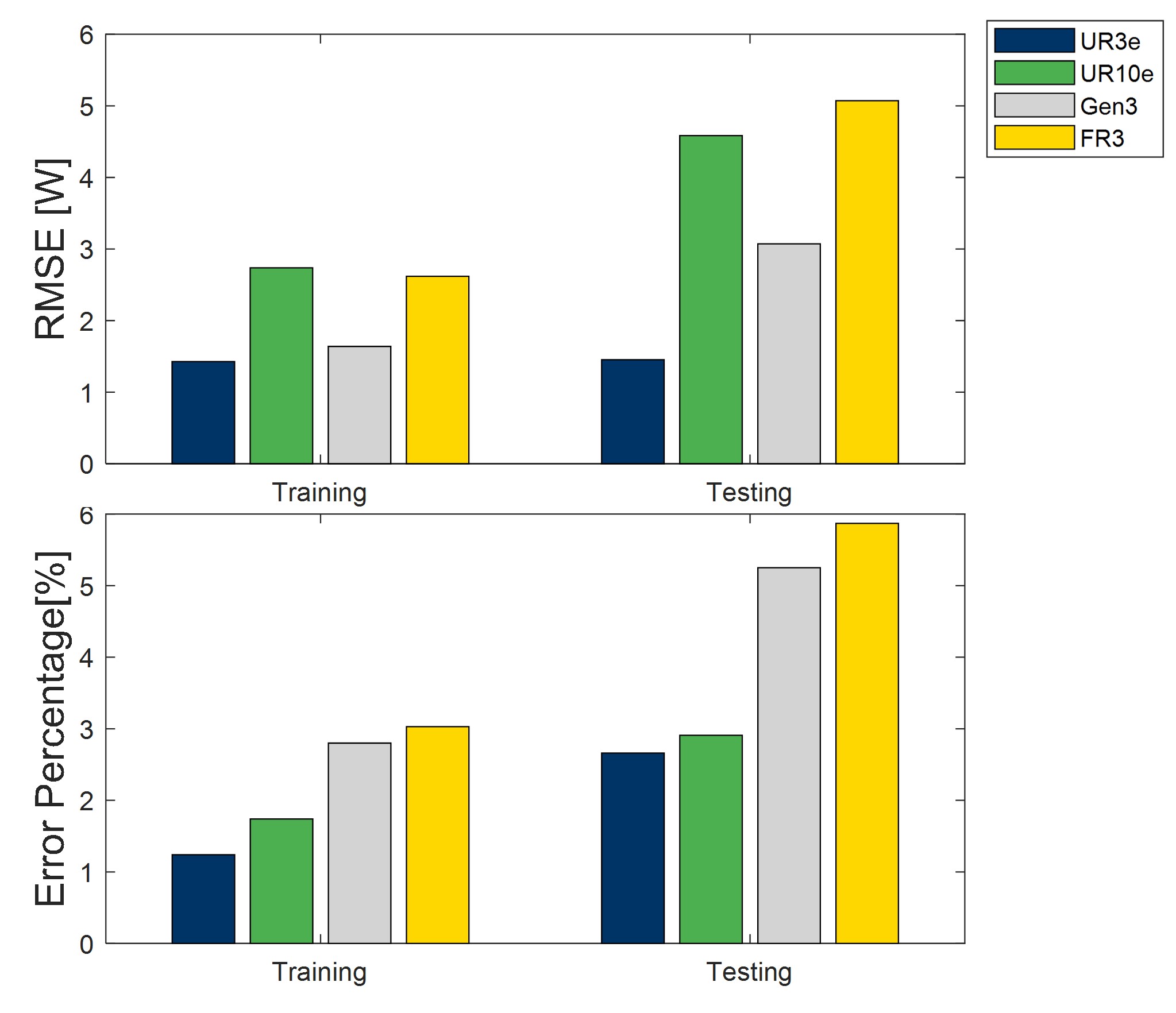}
  \vspace{-4mm}
  \caption{RMSE for training and testing dataset.}
  \label{fig:results_summary}
  \vspace{-6mm}
\end{figure}

We utilized the metrics defined in Equation \ref{RMSE} and Equation \ref{RMSEa} to evaluate the performance of the \ac{ec} model. The model's results are showcased in Table \ref{Tb:results}, which provides the RMSE, the relative percentage of RMSE, and $r^2$ for each robot. Fig. \ref{fig:results} displays a 60-second testing sample for each robot, while Fig. \ref{fig:results_summary} offers a summary of the RMSE across both the training and testing datasets.

In the case of the dynamic model, performance on the training dataset yielded an RMSE\% between 1.63 and 5.47, whereas for the testing dataset, the RMSE\% ranged from 2.52 to 9.00. As for the \ac{ec} model, the training dataset results exhibited an RMSE between 1.42 W and 2.80 W, while the testing dataset varied from 1.45 W to 5.25 W. The best model performance was achieved with the UR3e robot, while the Gen3e robot exhibited the least optimal performance.

A limitation of this methodology is the presupposition that the \ac{ec} of a robot's electronics — including its onboard computer, sensors, screen, and other general electronics — remains constant. This assumption's effect is mitigated in the case of Universal Robots by exclusively measuring the \ac{ec} of the manipulator, while disregarding controller consumption, using internal sensors. However, for the FR3 and Gen3 robots, the model's error is likely to be more pronounced. This discrepancy emerges because the datasets for these robots were collected using the external power meter HCM8050, which records power from both the controller and the manipulator. Consequently, the model's precision is compromised for these specific robots, as the RMSE percentage rises for the testing dataset.

\vspace{-1mm}
\section{Conclusions}
\vspace{-1mm}

This paper presents a novel methodology employed by the EcBot library to simulate the \ac{ec} of robotic manipulators. The library's approach is dependent on certain parameters provided by the robot manufacturers. These include the dimensions and weights of the links, Denavit-Hartenberg parameters, and various operational data. Given these inputs, the EcBot library can construct a model to estimate the energy usage of a given robotic manipulator. The library has been implemented and evaluated on four distinct robot models: the UR3e, UR10e, FR3, and Gen3. In order to validate the model, an array of tests was conducted. The results showed a root mean square error (RMSE) of 1.42 W to 2.80 W for the training dataset, and between 1.45 W and 5.25 W for the testing dataset.

However, the methodology does have a certain limitation. It assumes that the \ac{ec} of the robot's electronics remains constant. For the UR manipulators, the impact of this assumption is mitigated by utilizing the internal sensors that measure the electrical power of the manipulator. Conversely, for the FR3 and Gen3 robots, the power analyzer measures the total power of the manipulator and its controller, which could potentially lead to a larger assumption error. In future work, it will be necessary to refine the model to incorporate the \ac{ec} of the electronic components.



\bibliographystyle{IEEEtran}
\bibliography{references.bib}{}
\end{document}

%% file: acro.tex
\newacronym{ir}{IR}{Industrial Robot}
\newacronym{cobot}{Cobot}{Collaborative Robot}
\newacronym{ec}{EC}{Energy Consumption}
\newacronym{lwr}{LWR}{Leightweight Robot}